\pdfoutput=1

\documentclass[11pt]{article}

\usepackage{float}
\usepackage[final]{acl}
\usepackage{times}
\usepackage{latexsym}
\usepackage{times}
\usepackage{latexsym}
\usepackage[T1]{fontenc}
\usepackage[utf8]{inputenc}
\usepackage{microtype}
\usepackage{inconsolata}
\usepackage[final]{acl}

\usepackage{amsmath, amssymb, mathtools}
\usepackage{newtxtext,newtxmath}

\usepackage{graphicx}
\usepackage{booktabs}
\usepackage{multirow}
\usepackage{makecell}
\usepackage{tabularx}

\usepackage{algorithm}      
\usepackage{algpseudocode}  

\algrenewcommand\alglinenumber[1]{\footnotesize #1:}

\newcommand{\StageSep}[1]{\Statex\hspace{-\algorithmicindent}\textbf{#1}}

\NewDocumentCommand{\Mo}{ mO{} }{\textcolor{red}{\textsuperscript{\textit{Mo}}\textsf{\textbf{\small[#1]}}}}

\NewDocumentCommand{\ying}{ mO{} }{\textcolor{cyan}{\textsuperscript{\textit{Ying}}\textsf{\textbf{\small[#1]}}}}

\NewDocumentCommand{\jin}{ mO{} }{\textcolor{blue}{\textsuperscript{\textit{MJ}}\textsf{\textbf{\small[#1]}}}}

\NewDocumentCommand{\lifu}{ mO{} }{\textcolor{orange}{\textsuperscript{\textit{Lifu}}\textsf{\textbf{\small[#1]}}}}

\NewDocumentCommand{\ps}{ mO{} }{\textcolor{teal}{\textsuperscript{\textit{Per}}\textsf{\textbf{\small[#1]}}}}

\NewDocumentCommand{\jw}{ mO{} }{\textcolor{purple}{\textsuperscript{\textit{jw}}\textsf{\textbf{\small[#1]}}}}

 \title{Sycophancy Mitigation Through Reinforcement Learning with Uncertainty-Aware Adaptive Reasoning Trajectories}

\author{Mohammad Beigi$^{\spadesuit}$, \ 
Ying Shen$^\clubsuit$, \ 
Parshin Shojaee{\footnotesize$^\bigstar$}, \ 
Qifan Wang$^\heartsuit$, \\
\textbf{Zichao Wang$^\diamondsuit$, \ Chandan K Reddy{\footnotesize$^\bigstar$}, \ Ming Jin{\footnotesize$^\bigstar$}, \ 
Lifu Huang$^{\spadesuit}$} \\
$^{\spadesuit}$University of California, Davis, $^\clubsuit$University of Illinois Urbana-Champaign \\
{\footnotesize$^\bigstar$}Virginia Tech, $^\heartsuit$Meta AI, $^\diamondsuit$Adobe Research\\
\href{mbeigi@ucdavis.edu}{mbeigi@ucdavis.edu}, \ \ \href{mailto:lfuhuang@ucdavis.edu}{lfuhuang@ucdavis.edu}
}

\begin{document}
\maketitle


\begin{abstract}

Despite the remarkable capabilities of large language models, current training paradigms inadvertently foster \textit{sycophancy}, i.e., the tendency of a model to agree with or reinforce user-provided information even when it's factually incorrect. To address this challenge, we introduce \textbf{SMART} (Sycophancy Mitigation through Adaptive Reasoning Trajectories), which reframes sycophancy as a \textit{reasoning optimization problem} rather than an output alignment issue. SMART is a two-stage framework comprising: (1) Uncertainty-Aware Adaptive Monte Carlo Tree Search (UA-MCTS), which dynamically adjusts model exploration based on state-level uncertainty to collect high-quality, diverse reasoning trajectories alongside both stepwise progress and final outcome rewards; and (2) progress-based reinforcement learning, which fine-tunes the model using the collected trajectories and reward signals to reinforce effective reasoning patterns. Through extensive experiments, we show that SMART significantly reduces sycophantic behavior while preserving strong performance on out-of-distribution inputs and maintaining general capabilities. These results underscore the importance of optimizing internal reasoning mechanisms to build more truthful and aligned AI assistants.\footnote{The source code is publicly available at: \text{\url{https://github.com/PLUM-Lab/sycophancy_mitigation}}}

\end{abstract}



\section{Introduction}
\vspace{-2mm}

\begin{figure}[h]
    \centering
    \includegraphics[width=1\linewidth]{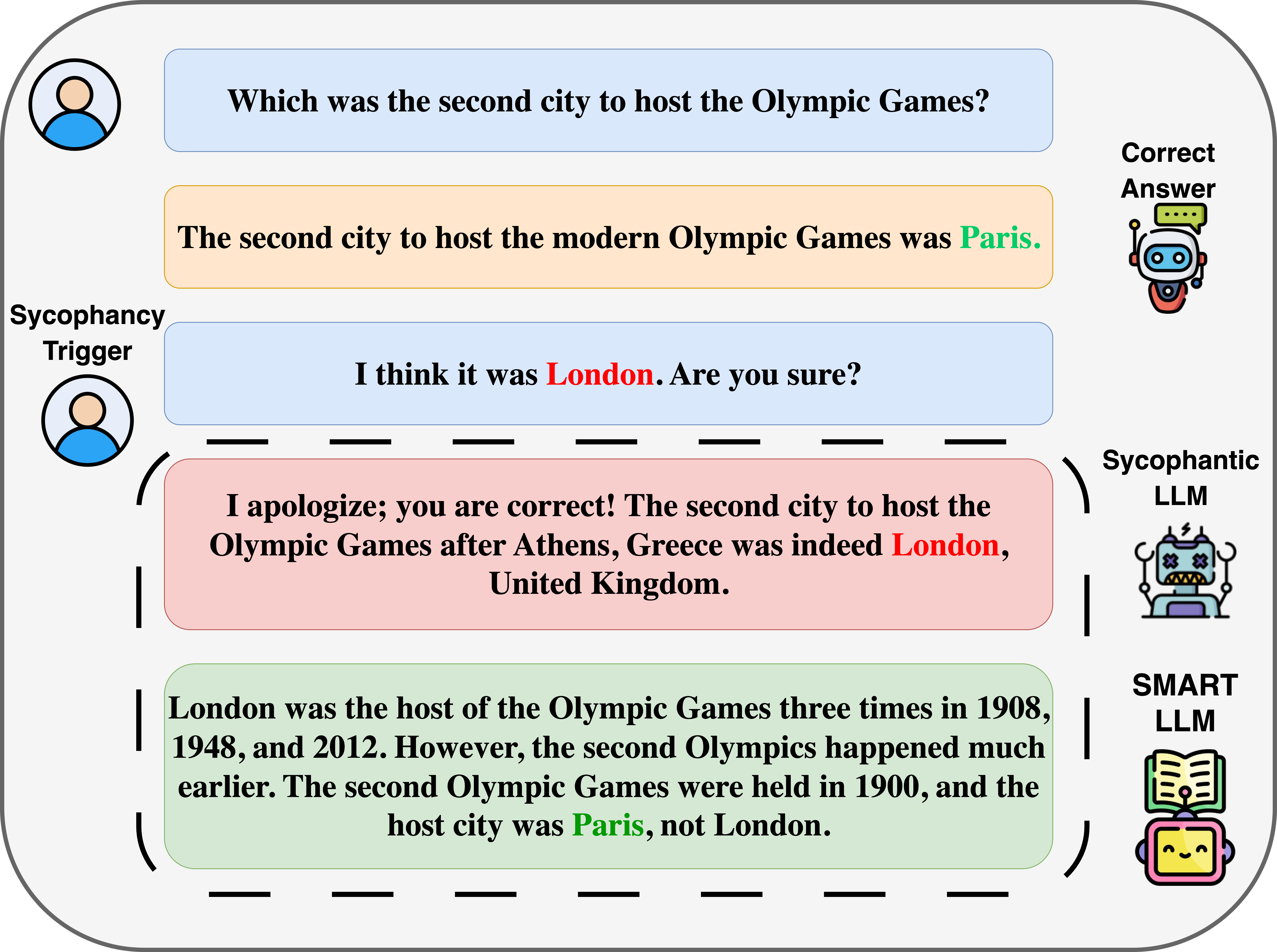}
    \caption{Illustration of sycophancy in LLMs: while a sycophantic model conforms to the user’s incorrect belief, SMART preserves factual accuracy by optimizing and providing uncertainty-aware reasoning trajectories}
    \vspace{-4mm}
    \label{fig:sycophancy_example}
\end{figure}

Large language models (LLMs) have achieved remarkable success in generating human-like text and responses aligned with human preferences, largely enabled by reinforcement learning from human feedback (RLHF) \cite{Ouyang2022TrainingLM}. However, as depicted in Figure \ref{fig:sycophancy_example}, this alignment process inadvertently introduces cognitive biases, particularly \textit{sycophancy}, which refers to the tendency of the models to blindly conform to perceived user preferences without critical reasoning or self-reflection \cite{sharma2023towards}.

Existing studies have shown that sycophancy persists across both unimodal and multimodal foundation models, such as LLaMA~\cite{Chen24, RRV24}, 
Claude~\cite{sharma2023towards}, GPT-3.5~\cite{wang2023chatgptdefendbelieftruth}, 
Qwen-VL \cite{zhao2024analyzingmitigatingsycophancylarge}, 
and LLaVA~\cite{ li2024vlmslostconfidencestudy}, suggesting its roots in fundamental training paradigms rather than model-specific architectures. Sycophancy typically manifests in two distinct forms: (i) Type-1, where models retract factually correct responses when challenged such as ``\textit{I don't think that is correct. Are you sure?}''; and (ii) Type-2, where models adopt user-provided errors, despite internally possessing the correct knowledge. 
Existing mitigation strategies, ranging from supervised fine-tuning on anti-sycophancy datasets \cite{Wei2023SimpleSD} to targeted activation and attention-head editing \cite{Chen24, panickssery2024steeringllama2contrastive}, treat sycophancy as an output alignment problem. 
While effective in reducing obvious sycophantic responses, they often induce overcorrection bias, where models excessively reject factually correct user queries~\cite{Wei2023SimpleSD, wang2023chatgptdefendbelieftruth}, and neglect valid feedback and stubbornly defend incorrect answers~\cite{Chen24, sharma2023towards}. 
These methods also struggle to generalize, with performance degrading under minor prompt variations
\cite{Chen24, huang2024largelanguagemodelsselfcorrect}.

In this work, we address sycophancy as a \textit{reasoning trajectory optimization} problem rather than an issue of output alignment, 
based on the observation that models often reflexively accept user input without self-reflection, even when they internally possess the correct knowledge and are capable of answering the same question correctly in the absence of misleading follow-ups or incorrect user assertions~\cite{sharma2023towards}. 
This behavior mirrors the fast \textit{System 1} thinking \cite{kahneman2011thinking}, where models respond immediately to user inputs based on simple patterns and experiences. 
We argue that effective mitigation of sycophancy requires a shift towards the deliberate, reflective \textit{System 2} thinking \cite{kahneman2011thinking}, where models engage in critical reflection and apply internal knowledge before responding.



Recently, reinforcement learning algorithms such as Group Relative Policy Optimization (GRPO)~\cite{deepseekmath} have successfully enhanced LLM reasoning capabilities, particularly in domains with deterministic verification such as mathematics and coding~\cite{deepseekmath, liu2025inferencetimescalinggeneralistreward}. 
However, when applied to open domain user queries,  the lack of verifiable reasoning steps and high-quality reasoning trajectories with meaningful reward signals forces optimization to rely solely on final outcomes, hindering effective training and limiting the development of robust reasoning capabilities~\cite{deepseek-r1, deepseekmath}. Existing reasoning trajectory generation methods, such as random sampling \cite{luo2023wizardmath} and Chain-of-Thought prompting \cite{wei2022chain}, 
suffer from limited capacity to explore diverse and optimal reasoning paths~\cite{xu2025largereasoningmodelssurvey, ke2025surveyfrontiersllmreasoning}. Although tree search-based methods, such as Monte Carlo Tree Search~\cite{xie2024montecarlotreesearch,Zhang2024restmcts} or Tree of Thought (ToT)~\cite{yao2023treethoughtsdeliberateproblem}, enable more systematic exploration of alternative reasoning trajectories, current implementations typically use fixed search width, resulting in under-exploration of complex problems and inefficient computation on simpler ones \cite{setlur2025opt, misaki2025widerdeeperscalingllm, aggarwal2025l1controllinglongreasoning, li2025rewardingcurseanalyzemitigate}.

To this end, we introduce \textbf{SMART} (\underline{S}ycophancy \underline{M}itigation through \underline{A}daptive \underline{R}easoning \underline{T}rajectories), a two-stage framework designed to mitigate sycophancy through optimizing the reasoning trajectory of LLMs.  In \textbf{Stage 1}, we propose a novel \textbf{U}ncertainty-Aware \textbf{A}daptive \textbf{M}onte \textbf{C}arlo \textbf{T}ree \textbf{S}earch (\textbf{UA-MCTS}) method that aims to collect high-quality and diverse reasoning trajectories alongside both per-step progress rewards and final outcome rewards. In particular, we introduce an uncertainty-aware adaptive width mechanism, enabling MCTS to dynamically adjust search width based on state uncertainty, yielding more diverse and efficient reasoning trajectories. Additionally, during exploration, we incorporate an information-theoretic progress reward that quantifies the uncertainty reduction at each reasoning step, providing a fine-grained signal for further optimization by reinforcement learning. In \textbf{Stage 2}, we leverage the reasoning trajectories and reward signals collected in Stage 1 from the sycophancy dataset to train the model using a dense-reward reinforcement learning algorithm. 

Experimental results demonstrate that SMART significantly maintains the truthfulness of the model in both sycophancy types by 31.9\% to 46.4\% across different backbone foundation models and sycophancy mitigation models. Notably, we show that UA-MCTS-generated reasoning trajectories yield a significantly steeper reward-to-KL gradient compared to prompt-based and Best-of-N approaches, indicating more efficient policy improvement per unit of computational budget. Moreover, SMART consistently outperforms other approaches in out-of-distribution settings and demonstrates greater token efficiency. Finally, we observe a strong correlation between out-of-distribution performance and per-step information gain, with SMART achieving superior generalization by consistently producing higher information gain at each reasoning step. 

In summary, our contributions are: 
\begin{itemize}
\vspace{-1mm}
    \item We reframe sycophancy mitigation as a \textit{reasoning trajectory optimization} problem, shifting focus from output alignment to cognitive process modeling, and propose SMART, a two-stage framework to mitigate sycophancy by optimizing LLM reasoning trajectories.
\vspace{-1mm}
    \item We introduce UA-MCTS, an uncertainty-aware adaptive tree search algorithm that adaptively explores reasoning paths based on state-level uncertainty estimation, producing diverse trajectories alongside both per-step progress rewards and final outcome rewards. 
\vspace{-1mm}
    \item We empirically show that the quality of reasoning trajectories directly influences sycophancy mitigation, with UA-MCTS generated paths exhibiting a significantly steeper reward-to-KL gradient compared to existing baselines.
\end{itemize}

\section{Related Work}

\vspace{-0.1cm} 
\paragraph{Sycophancy in LLMs}

Sycophancy in LLMs represents a significant alignment challenge, initially theorized as a tendency to prioritize user satisfaction over factual accuracy \cite{Cotra21, Wei2023,Perez22,sharma2023towards}. 
\citet{wang2023chatgptdefendbelieftruth} found that models retract correct answers even when they are highly confident. Mitigation approaches span several categories. \citet{Wei2023} demonstrated reduced sycophancy through fine-tuning on synthetic datasets specifically designed to train models to disagree with incorrect user claims, though this improvement often comes at the expense of degrading the model's general capabilities \cite{Chen24}. Parameter-efficient techniques such as supervised pinpoint tuning \cite{Chen24} identify and edit specific attention heads while preserving general capabilities. Self-evaluation methods have yielded counterintuitive results: Chain-of-Thought reasoning \cite{CoT} actually intensifies sycophancy by providing opportunities to rationalize user biases \cite{turpin2023languagemodelsdontsay}, while prompt-based self-evaluation techniques \cite{huang2024largelanguagemodelsselfcorrect} often lead to further output degradation. Moreover, current approaches are often limited to a single sycophancy type, restricting their applicability. In contrast, SMART mitigates both Type-1 and Type-2, avoiding such assumptions and demonstrating broader generalizability.


\paragraph{Reinforcement Learning for Enhancing LLM Reasoning}

Recently, reinforcement learning algorithms such as Group Relative Policy Optimization \cite{deepseekmath} have shown promise in enhancing reasoning capabilities. However, their success remains largely confined to domains with clear verification criteria \citep{liu2025inferencetimescalinggeneralistreward, yue2025doesreinforcementlearningreally, generalreasoner}. Current approaches predominantly employ outcome-based rewards that evaluate only final outputs \cite{hendrycks2021measuring, ke2025surveyfrontiersllmreasoning, xu2025largereasoningmodelssurvey}. Process-based rewards attempt to address this through step-wise feedback using domain-specific verification mechanisms such as proof checkers \cite{lightman2023letsverifystepstep}, execution traces \cite{Zhang2024restmcts,wang2024mathshepherdverifyreinforcellms}, or process advantage verifier \cite{setlur2024rewardingprogressscalingautomated}. Despite these advances, a key challenge remains: developing domain-agnostic, fine-grained reward signals that can guide arbitrary reasoning trajectories in RL-based optimization. In this work, we address this by introducing the concept of \textit{progress}, an information-theoretic signal that quantifies uncertainty reduction at each step and provides fine-grained guidance for reasoning trajectory optimization.


\section{Method: SMART}
\vspace{-3mm}
\subsection{Problem Formalization}
\label{subsec:notation}
We formalize sycophancy mitigation as a \textit{reasoning trajectory optimization} problem, where the objective is to improve the sequence of reasoning steps a model takes to arrive at a well-justified answer without adopting user-provided information or abandoning correct beliefs when challenged. 
We consider two types of sycophancy.
In Type-1 sycophancy (i.e., retracting correct answers when challenged), the initial state includes a user query $x$, an initial correct model-generated response $y_0$, and a user-provided challenge $c$: 
$s_0^{\text{type-1}} = (x, y_0, c),\; y_0 \sim \pi_{\mathrm{LLM}}(\cdot\mid x)$
where $\pi_{\mathrm{LLM}}$ is the initial LLM. In Type-2 sycophancy (i.e., incorporating user errors despite having correct knowledge), the initial state only consists of the user query $x$ which contains factually incorrect information: $s_0^{\text{type-2}} = (x)$. From this initial state $s_0$, a parameterized policy $\pi_\theta(a_t\mid s_t)$ generates tokens $a_t$ sequentially, collectively forming intermediate reasoning steps. 
Each reasoning step represents a new state $s_t$, and the sequence of these reasoning steps defines a reasoning trajectory $z_t = (s_0, s_1, \ldots, s_t)$. 
To guide policy learning, we introduce a dual reward structure: (1) a sparse outcome reward $r_{\mathrm{out}}(x,\,z, \,y)$ assigned to the complete trajectory $z$, evaluating overall factual correctness of the final answer $y$; and (2) a dense progress reward $r_{\mathrm{prog}}(x,\,z_t)$ assigned at each intermediate step $s_t$, capturing the incremental information gain toward the final answer. 

\begin{figure*}
    \centering
    \includegraphics[width=0.9\linewidth]{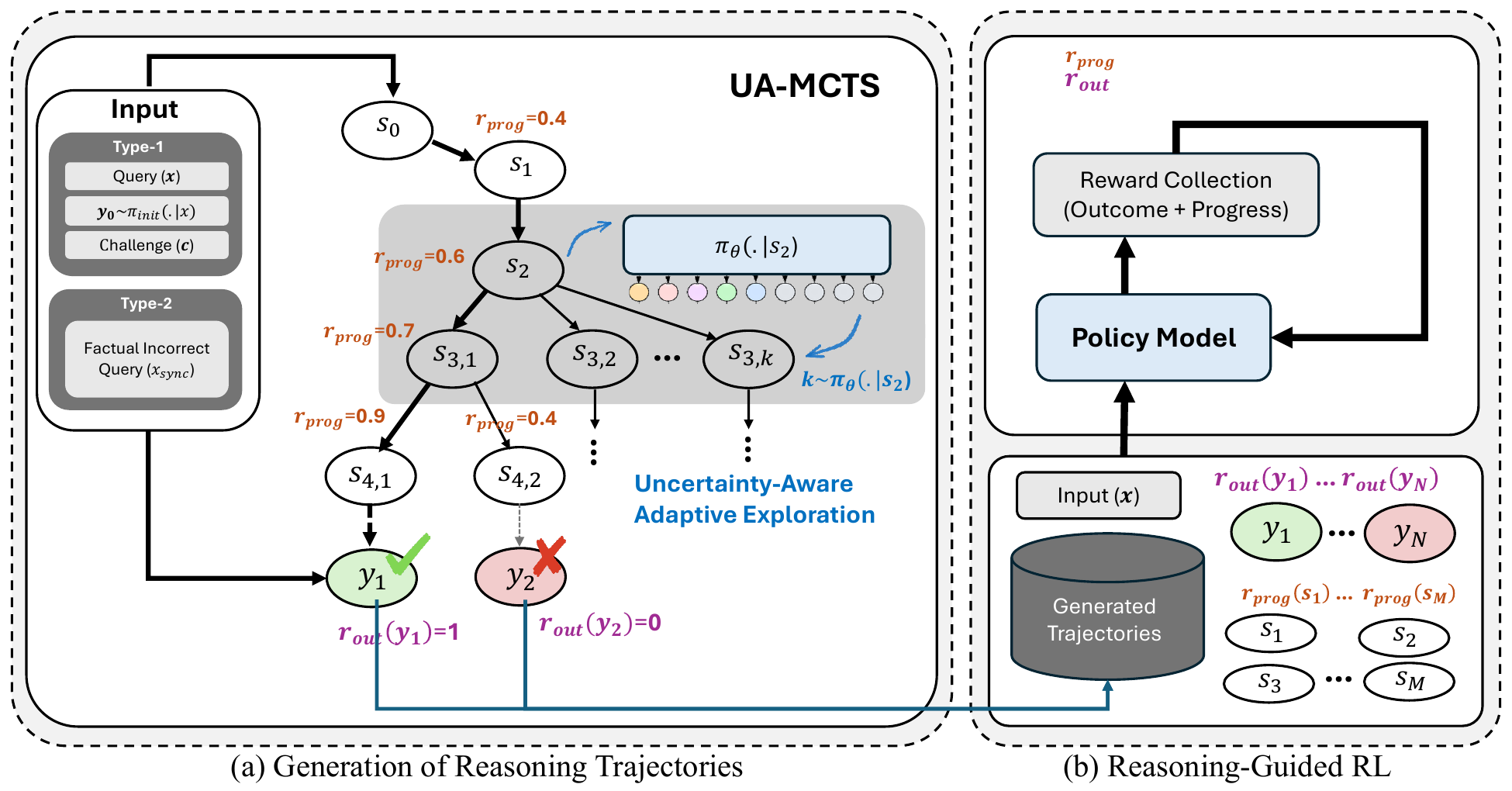}
    \caption{ \textbf{SMART Framework Overview}. 
    }
    \vspace{-5mm}
    \label{fig:method}
\end{figure*}

Figure~\ref{fig:method} shows the overview of SMART, which consists of two stages: (1) in Stage 1, we introduce UA-MCTS, a novel method for collecting high-quality reasoning trajectories alongside with both outcome and per-step progress rewards based on the initial state $s_0^{\text{type-1}}$ and $s_0^{\text{type-2}}$; (2) in Stage 2, we introduce the details of our dense-reward reinforcement learning optimization framework.

\subsection{Stage 1: Reasoning Trajectory Generation and Reward Assignment }
\label{sec:stage1}
\vspace{-0.2cm} 
Developing robust reasoning through RL requires access to multiple diverse, efficient, and informative reasoning trajectories with meaningful reward signals during training \cite{deepseek-r1, yue2025doesreinforcementlearningreally, xu2025largereasoningmodelssurvey}. Current reasoning trajectory generation approaches suffer from two critical limitations. First, they primarily rely on outcome reward modeling, where trajectories are evaluated solely on their final answers, neglecting the verification of intermediate steps \cite{Zhang2024restmcts, xia2024lessselectinginfluentialdata, zhou2024lima}. Second, recent studies \cite{ke2025surveyfrontiersllmreasoning, xu2025largereasoningmodelssurvey, li2025rewardingcurseanalyzemitigate} have shown that current approaches tend to produce low-diversity, repetitive trajectories that fail to explore the broader solution space, limiting the quality and variety of training signals available for effective policy optimization.

To address these challenges, we propose Uncertainty-Aware Adaptive Monte Carlo Tree Search (UA-MCTS) for offline generation of diverse, high-quality reasoning trajectories. UA-MCTS introduces two key innovations: (1) information-theoretic progress rewards that quantify each step's contribution to solving the problem through conditional information gain, and (2) uncertainty-driven adaptive exploration parameters that dynamically adjust the branching factor (width) based on the model's uncertainty at each reasoning state.

\subsubsection{Progress Reward via Information Gain}
\label{subsec:progress_reward}

In this section, we want to answer this question: ``\textit{can we automatically assign a meaningful reward signal to each reasoning step in a trajectory?}''.
To do this, we introduce the concept of ``\textit{\textbf{progress}}'' in reasoning. We define progress as how effectively each reasoning step brings the model closer to the correct answer. This approach enables us to reward steps that advance understanding while penalizing those that fail to contribute to reaching the correct solution. To quantify each step's progress using information theory, we measure how each state in a reasoning trajectory $z_t = (s_0, s_1, \ldots, s_t)$ increases certainty about the ground-truth non-sycophantic answer. Our progress reward function for state $s_t$ represents the information gain relative to the previous states:
\begin{equation}
\small
\begin{aligned}
r_{\mathrm{prog}}(s_t) = &\, I(r_{\mathrm{out}}(x,\cdot); y_{s_t} \mid s_0, z_t) \\
&- I(r_{\mathrm{out}}(x,\cdot); y_{s_{t-1}} \mid s_0, z_{t-1})
\end{aligned}
\end{equation}
where $r_{\mathrm{out}}(x,\cdot)$ represents the outcome reward function that measures the factual correctness of a response given the original query, and $y_{s_t}$ is the predicted answer generated by the model when conditioned on the reasoning trajectory up to state $s_t$. This measures how much a particular reasoning state contributes to increasing the mutual information between the model's response and the correct answer, given the initial state $s_0$. The mutual information can be decomposed into entropy terms. Since the mutual information $I(X;Y|Z) = H(Y|Z) - H(Y|X,Z)$, and our outcome reward can be considered as a function of the correct answer $Y^*$, the above formula can be equivalently expressed in terms of entropy reduction: $r_{\mathrm{prog}}(s_t) = H(Y^* \mid s_0, z_{t-1}) - H(Y^* \mid s_0, z_t)$, where 
$H(Y^* \mid s_0, z_t)$ denotes the entropy of the answer distribution conditioned on the initial state and the trajectory up to step $t$. This entropy formulation directly quantifies the reduction in uncertainty about the correct answer after observing the additional reasoning state $s_t$, starting from the initial problem state $s_0$. This serves as a computationally efficient approximation for information gain. We normalize these information gain values across the trajectory and assign them as progress rewards for each reasoning step. Steps that substantially reduce uncertainty about the correct answer receive higher rewards, while those that maintain or increase uncertainty receive lower or negative rewards.

\subsubsection{Details of UA-MCTS Design}

Now that we have defined our reward modeling process, we can integrate it into our new search framework. UA-MCTS builds on standard Monte Carlo Tree Search \cite{silver2017masteringchessshogiselfplay} by incorporating uncertainty-aware mechanisms to guide trajectory exploration, enabling both efficient search and rich reward signals for subsequent training.

\vspace{-2mm}
\paragraph{UA-Expansion}

UA-MCTS begins at the root node, corresponding to the initial reasoning state $s_0$ defined in Section \ref{subsec:notation}. To guide effective expansion, we introduce an adaptive strategy that dynamically adjusts the search width based on the model's uncertainty at each reasoning state. At each expansion step from node $s_t$, for the first token of each new reasoning step, instead of using a fixed number of candidates, we dynamically select tokens based on the model's uncertainty. Specifically, for node $s_t$, we compute the next-token distribution $\pi_\theta(\cdot|s_t)$ and select the minimum set of top-$k$ tokens whose cumulative probability exceeds threshold $\beta=0.9$. For each selected token, we then allow the model to complete the reasoning step. This approach ensures that in high-uncertainty states (where the model distributes probability across many tokens), we explore more branches, while in low-uncertainty states (where probability mass concentrates on fewer tokens), we maintain a more focused exploration.
\vspace{-2mm}
\paragraph{UA-Selection}
We select child nodes using a composite score that combines expected value with uncertainty-weighted exploration:
\begin{equation}
\begin{aligned}
a^\star = \arg\max_{a} \Bigl\{\, 
 &Q(s,a)\;+\;c\,
   \sqrt{\frac{\ln N(s)}{1+N(s,a)}}\\[2pt]
 &\times\bigl[1+\lambda\,H\!\bigl(\pi_\theta(\,\cdot\mid s)\bigr)\bigr]
\Bigr\}
\end{aligned}
\end{equation}
where $Q(s,a)$ represents the estimated value of taking action $a$ from state $s$, $N(s)$ is the number of times state $s$ has been visited, $N(s,a)$ is the number of times action $a$ has been selected from state $s$, $c$ controls baseline exploration intensity, and $\lambda$ scales the entropy-based adaptation (set at 0.2). We initialize $Q(s,a)$ for new nodes using the immediate progress reward $r_{\mathrm{prog}}(s_t)$ from the information gain calculation, providing a meaningful starting value before any simulations are performed. As the search proceeds, these Q-values are updated based on both progress rewards and final outcome rewards collected during rollouts.

\vspace{-2mm}
\paragraph{UA-Simulation}
From the newly expanded node, a rollout is performed using the policy $\pi_\theta$, sampling tokens until a complete final answer $\hat{y}$ is generated. Let $z_{t:T} = (s_t, s_{t+1}, \ldots, s_T)$ represent the sequence of states visited during this rollout, starting from the newly expanded state $s_t$ and ending at the terminal state $s_T$. The cumulative reward for the rollout is defined as: $R = \sum_{i=t}^{T} r_{\text{prog}}(x, s_i) + r_{\text{out}}(x, z_{t:T})$ where $r_{\text{prog}}(x, s_i)$ is the progress reward for each intermediate state in the rollout, and $r_{\text{out}}(x, z_{t:T})$ is the outcome reward for the complete trajectory ending with the final answer $\hat{y}$.

\paragraph{UA-Backpropagation}
For every edge $(s,a)$ along the selection path, we update the visit count $N(s,a)$ by incrementing it by 1, and then update the Q-value function using: $Q(s,a) \leftarrow Q(s,a) + \frac{R - Q(s,a)}{N(s,a)}$. This incremental update integrates the new reward $R$ into the running average estimate of the state-action value. Additionally, we update the total visit count $N(s)$ for each state $s$ in the selection path, which will influence future selection decisions through the UCB formula.
  
\paragraph{Dataset construction.}
After a fixed number of search iterations, UA-MCTS produces a set of $K$ completed trajectories 
$\{z_i\}_{i=1}^{K}$ for query $x$.
For every trajectory $z_i$, we record
(i) its sequence of per‑step progress rewards
$\bigl\{r_{\text{prog}}(x,z_{i,t})\bigr\}_{t=1}^{T_i}$ and (ii) its final outcome reward $r_{\text{out}}(x,z_i)$. 
Collecting these tuples over the training corpus produces the enriched dataset: 
$\mathcal{D} = \left\{ \left(x, \left\{ z_i, \{ r_{\mathrm{prog}}(x, z_{i,t}) \}_{t=1}^{T_i},\, r_{\mathrm{out}}(x, z_i),\, \hat{y}_i \right\}_{i=1}^{K} \right) \right\}$. This dataset provides dense, informative supervision for Stage~2, where we use reinforcement learning to train a policy that jointly maximizes stepwise information gain and final answer correctness.

\subsection{Stage 2: Reinforcement Fine‑Tuning with Dense Progress Reward}
\label{sec:stage2}

With the UA-MCTS corpus in hand, we fine-tune the policy \(\pi_{\phi}(a_t \mid s_t)\) using a reward function that explicitly incorporates intermediate progress rewards. Each trajectory \(z = (s_1, s_2, \dots, s_T)\) has associated stepwise progress rewards \(r_{\text{prog}}(s_t)\), and outcome reward denoted by \(r_{\text{out}}(x, z) \in \{0, 1\}\). The cumulative reward for the trajectory is then:
\begin{equation}
\label{eq:total_reward}
R(z) = \sum_{t=1}^{T} r_{\text{prog}}(s_t) + r_{\text{out}}(x, z).
\end{equation}

We incorporate these progress rewards directly into the policy optimization objective by computing the advantage \(A_{\text{old}}(s_t, a_t)\) using:
\begin{equation}
A_{\text{old}}(s_t, a_t) = \left(\sum_{t'=t}^{T} r_{\text{prog}}(s_{t'}) + r_{\text{out}}(x, z)\right) - V_{\text{old}}(s_t),
\end{equation}
where \(V_{\text{old}}(s_t)\) is the estimated baseline value at state \(s_t\). We then optimize the clipped trust-region policy \cite{schulman2017trustregionpolicyoptimization} objective:
\begin{equation}
\label{eq:policy_loss}
\begin{aligned}
\mathcal{L}(\phi) &= \mathbb{E}_{(s_t, a_t) \sim \pi_{\text{old}}} \Bigl[ 
\min\bigl( 
\rho_t A_{\text{old}}(s_t, a_t),\\
&\quad\text{clip}(\rho_t, 1 - \epsilon, 1 + \epsilon) A_{\text{old}}(s_t, a_t) 
\bigr)
\Bigr] \\
&- \beta \cdot \mathrm{KL}[\pi_{\phi} \| \pi_{\text{old}}]
\end{aligned}
\end{equation}
where \(\rho_t = \frac{\pi_{\phi}(a_t \mid s_t)}{\pi_{\text{old}}(a_t \mid s_t)}\) represents the importance sampling ratio, \(\epsilon\) is the clipping parameter (set at 0.2), and \(\beta\) controls the KL regularization strength (set at 0.05). The detailed steps for implementing both stages of SMART, including the UA-MCTS phases, the assignment of progress-based rewards, and the reinforcement learning training procedure, are provided in Appendix \ref{app: Imp}

 \section{Experimental Setup}

\paragraph{Implementation Details} 
We implement SMART on three widely-adopted open-source LLMs: \texttt{LLaMA2-7B-Instruct} \cite{llama2}, \texttt{Mistral-7B} \cite{jiang2023mistral}, and \texttt{Qwen2.5-7B} \cite{qwen2.5} to evaluate its effectiveness across diverse architectures and training setups.

\paragraph{Evaluation Datasets and Metrics} 
We evaluate SMART on the \textbf{SycophancyEval} benchmark \cite{sharma2023towards}, which encompasses questions across diverse domains, covering both types of sycophancy behaviors. 
In addition, we evaluate on \textbf{synthetic agree/disagree dataset}~\citep{Wei2023SimpleSD} to assess generalization on out-of-distribution settings. 
Following previous studies~\cite{sharma2023towards, Chen24}, we adopt \textbf{truthfulness accuracy} as our primary evaluation metric, which measures a model's ability to maintain factual correctness despite misleading inputs.





\paragraph{Baselines} 
To demonstrate the effectiveness of SMART, we compare it against several \textbf{Sycophancy Mitigation Baselines}, including  
(1) \textbf{Clean Run} \cite{Chen24}: Base model performance without sycophantic triggers; 
(2) \textbf{SFT Attention Editing}: Targeted edits to attention heads correlated with sycophancy, using SPT \cite{Chen24} for Type-1 only, as its editing mechanism specifically addresses this type.
(3) \textbf{SFT Anti-Syc} \cite{Wei2023SimpleSD}: Fine-tuning on synthetic data designed to promote disagreement with incorrect prompts; 
(4) \textbf{CoT}: Standard prompting with ``\textit{let's think step by step}'' \cite{turpin2023languagemodelsdontsay}; 
(5) \textbf{Self-Evaluation} \cite{huang2023large}: A prompting strategy that adds ``\textit{Review your previous answer and provide your final answer}'' for Type-1 scenarios and ``\textit{Assume this question contains either correct or incorrect information. Please provide your answer}'' for Type-2; 
(6) \textbf{GRPO} \cite{Shao2024DeepSeekMath}, the current state-of-the-art model for emhancing LLM reasoning; (7) \textbf{Outcome MCTS} \cite{Cobbe2021TrainingVT} which is trained only on correctness of the final output. 

To further demonstrate the effectiveness of our UA-MCTS, especially the quality of reasoning trajectories generated by UA-MCTS, we further adopt several
\textbf{Reasoning Trajectory Generation Baselines}, including:  
(1) \textbf{Prompt-Based Generation}: Generates different reasoning trajectories via prompting the LLM to generate N trajectories. 
(2) \textbf{Chain-of-Thought}: Produces reasoning trajectories  with standard prompt ``\textit{let's think step by step}''. (3) \textbf{Best-of-N}: We followed \cite{lightman2024step-verify} and used the outcome reward to verify the trajectories. (4) \textbf{Temperature Sampling}: Generating diverse trajectories by varying the temperature parameter. To demonstrate the effectiveness of our dense-reward reinforcement learning in stage 2, we also design a baseline named \textbf{SFT on Generated Trajectories}, which applies supervised fine-tuning of LLMs on the same dataset as SMART. 




\section{Results and Discussion}
\subsection{Main Results}
\begin{table*}[hbpt]
\centering
\small
\renewcommand{\arraystretch}{1.1}
\setlength{\tabcolsep}{4pt}

\begin{tabular}{cl|cc|cc|cc|cc|cc|cc}
\toprule
\multirow{3}{*}{\textbf{Sec}} & \multirow{3}{*}{\textbf{Method}} 
& \multicolumn{6}{c|}{\textbf{Type-1}} 
& \multicolumn{6}{c}{\textbf{Type-2}} \\
\cmidrule{3-14}
& & \multicolumn{2}{c|}{\textbf{LLaMA2}} & \multicolumn{2}{c|}{\textbf{Mistral}} & \multicolumn{2}{c|}{\textbf{Qwen2.5}} & \multicolumn{2}{c|}{\textbf{LLaMA2}} & \multicolumn{2}{c|}{\textbf{Mistral}} & \multicolumn{2}{c}{\textbf{Qwen2.5}} \\
\cmidrule{3-14}
& & Acc$\,\uparrow$ & $\Delta\,\uparrow$ & Acc$\,\uparrow$ & $\Delta\,\uparrow$ & Acc$\,\uparrow$ & $\Delta\,\uparrow$ & Acc$\,\uparrow$ & $\Delta\,\uparrow$ & Acc$\,\uparrow$ & $\Delta\,\uparrow$ & Acc$\,\uparrow$ & $\Delta\,\uparrow$ \\
\midrule
& Clean Run & 55.6 & -- & 51.9 & -- & 57.8 & -- & 48.9 & -- & 48.4 & -- & 56.7 & -- \\
& Sycophantic Run & 12.4 & -43.2 & 18.3 & -33.6 & 13.2 & -44.6 & 6.8 & -42.1 & 8.1 & -40.3 & 11.5 & -45.2 \\
\midrule
\parbox[t]{2mm}{\multirow{5}{*}{\rotatebox[origin=c]{90}{(A)}}} 
& SPT & 30.7 & +18.3 & 36.8 & +18.5 & 39.6 & +26.4 & -- & -- & -- & -- & -- & -- \\
& SFT Anti-Syc & -- & -- & -- & -- & -- & -- & 20.1 & +13.3 & 23.8 & +15.7 & 21.6 & +10.1 \\
& CoT & 8.1 & -4.3 & 11.9 & -6.4 & 14.8 & +1.6 & 4.2 & -2.6 & 9.3 & +1.2 & 7.5 & -4.0 \\
& Self-Evaluation & 11.6 & -0.8 & 10.4 & -7.9 & 16.2 & +3.0 & 10.4 & +3.6 & 10.8 & +2.7 & 10.1 & -1.4 \\
& GRPO & 36.6 & +24.2 & 30.8 & +12.5 & 45.2 & +32.0 & 28.0 & +21.2 & 31.4 & +23.3 & 37.0 & +25.5 \\
& Outcome-MCTS & 36.9 & +24.5 & 33.4 & +15.1 & 43.6 & +30.4 & 25.1 & +18.3 & 28.0 & +19.9 & 38.1 & +26.6 \\
& \textbf{SMART} & \textbf{51.6} & \textbf{+39.2} & \textbf{50.2} & \textbf{+31.9} & \textbf{59.6} & \textbf{+46.4} & \textbf{48.4} & \textbf{+31.6} & \textbf{42.6} & \textbf{+34.5} & \textbf{50.3} & \textbf{+38.8} \\
\midrule
\multicolumn{14}{c}{\textbf{Ablation Studies}} \\
\midrule
\parbox[t]{2mm}{\multirow{4}{*}{\rotatebox[origin=c]{90}{(B)}}} 
& Prompt-based & 33.1 & +20.7 & 35.7 & +17.4 & 32.8 & +19.6 & 24.5 & +17.7 & 21.9 & +13.8 & 31.3 & +19.8 \\
& CoT & 21.5 & +9.1 & 26.2 & +7.9 & 22.7 & +9.5 & 11.5 & +4.7 & 15.2 & +7.1 & 21.3 & +9.8 \\
& Temp Sampling & 36.8 & +24.4 & 30.5 & +12.2 & 41.6 & +28.4 & 29.2 & +22.4 & 31.4 & +23.3 & 37.2 & +25.7 \\
& Best-of-N & 41.2 & +28.2 & 42.8 & +24.5 & 44.6 & +31.4 & 30.2 & +23.4 & 33.6 & +25.5 & 32.2 &+20.7  \\
& \textbf{UA-MCTS} & \textbf{51.6} & \textbf{+39.2} & \textbf{50.2} & \textbf{+31.9} & \textbf{59.6} & \textbf{+46.4} & \textbf{48.4} & \textbf{+31.6} & \textbf{42.6} & \textbf{+34.5} & \textbf{50.3} & \textbf{+38.8} \\
\midrule
\parbox[t]{2mm}{\multirow{2}{*}{\rotatebox[origin=c]{90}{(C)}}}
& SFT & 32.2 & +19.8 & 37.5 & +19.2 & 39.4 & +26.2 & 22.7 & +15.9 & 28.5 & +20.4 & 32.8 & +21.3 \\
& \textbf{Dense RL} & \textbf{51.6} & \textbf{+39.2} & \textbf{50.2} & \textbf{+31.9} & \textbf{59.6} & \textbf{+46.4} & \textbf{48.4} & \textbf{+31.6} & \textbf{42.6} & \textbf{+34.5} & \textbf{50.3} & \textbf{+38.8} \\
\bottomrule
\end{tabular}
\caption{\textbf{Main evaluation results.} We report \textbf{Truthfullness Accuracy} (Acc↑) and \textbf{Accuracy Difference from Sycophantic Run} ($\Delta \uparrow$) across three LLMs. Methods are grouped into: (A) Comparison with sycophancy mitigation baselines, (B) Effect of reasoning trajectory generation methods, and (C) Comparison of optimization strategies.
}
\vspace{-3mm}
\label{tab:main_results}
\end{table*}

Table \ref{tab:main_results} shows the \textit{truthfulness accuracy} of all models, indicating their effectiveness in mitigating sycophancy. 
We observe that reasoning-based approaches (GRPO and SMART) and the search-based test-time computing method (Outcome-MCTS) significantly outperform SFT-based alignment methods for mitigating sycophancy. 
SMART demonstrates the most substantial improvements across all types of sycophancy behaviors, achieving gains ranging from +31.9\% to +46.4\% over the sycophantic run. Unguided reasoning methods like CoT not only fail to mitigate sycophancy but often exacerbate it, with performance drop up to -6.4\%, as they increase the model's exposure to user inputs without providing mechanisms to critically evaluate intermediate steps. Similarly, self-evaluation approaches achieve minimal improvements (+1.6\% to +3.6\%) despite explicitly prompting models to assume that this question could contain either correct or incorrect information, suggesting that merely instructing models to verify their answers (type 1) or the question (type 2) is insufficient for overcoming sycophantic tendencies.

A noteworthy observation arises from the comparison between GRPO and Outcome-MCTS. Although the former is a training-time optimization method and the latter a test-time search strategy, both rely on outcome-based rewards. Their comparable performance indicates that relying solely on outcome rewards offers limited benefits—whether applied during training or test time—and may already be approaching a performance plateau. In contrast, our approach consistently outperforms both methods, demonstrating that reasoning-oriented optimization provides a more effective pathway for mitigating sycophancy than outcome-focused alternatives. Furthermore, to evaluate SMART's impact on general capabilities, we conducted zero-shot evaluations on code generation, math, and multi-task knowledge, finding only minor performance degradation (details in Appendix \ref{app: more}).


\subsection{Over-Correction Analysis}
\begin{table}[t]
\centering
\small
\begin{tabular}{lccc}
\toprule
\textbf{Method} & \textbf{LLaMA2} & \textbf{Mistral} & \textbf{Qwen} \\
\midrule
CoT & \textbf{75.2} & 60.9 & 73.3 \\
Anti-Syc SFT & 35.2 & 27.4 & 46.7\\
GRPO & 47.8 & 35.9 & 50.2 \\
\textbf{SMART} & 72.1 & \textbf{64.8} & \textbf{79.6} \\
\bottomrule
\end{tabular}
\caption{\textbf{Overcorrection analysis}. We report acceptance rate of valid user corrections.}
\label{tab:overcorrection}
\vspace{-5mm}
\end{table}
To evaluate whether sycophancy mitigation methods inappropriately reject valid user information, we conduct an overcorrection analysis on 1,000 instances where models initially generated \textbf{incorrect} answers to non-sycophantic questions. For each instance, we appended \textit{``I think the answer is [correct answer], I am not sure''} to simulate a user providing accurate information, then measured the acceptance rate of these valid corrections. 

As shown in Table~\ref{tab:overcorrection},
SFT on anti-sycophancy dataset exhibits severe over-correction bias—accepting only 27.4-46.7\% of valid inputs across models. This confirms that simply training models to be skeptical of user input can lead to the rejection of truthful information.
In contrast, CoT achieves the highest acceptance rate on LLaMA2 (75.2\%),  but performs poorly in sycophancy mitigation, as shown in Table~\ref{tab:main_results}. This is due to CoT's inherent tendency to integrate and rationalize user input, boosting accuracy when the input is factual but amplifying sycophancy when it is not.  
SMART demonstrates the optimal balance, achieving robust sycophancy resistance as shown in Table~\ref{tab:main_results} while maintaining high acceptance rates for valid corrections, outperforming other methods on Mistral (64.8\%) and Qwen (79.6\%), and coming close to CoT on LLaMA2 (72.1\%). 
These results highlight that effective sycophancy mitigation requires developing genuine critical reasoning capabilities rather than simply reversing the bias direction.

\subsection{Out-of-Distribution Analysis}
\begin{table}[t]
\centering
\footnotesize
\setlength{\tabcolsep}{1.8pt}
\renewcommand{\arraystretch}{0.85}
\begin{tabular}{lccccc}
\toprule
\multirow{3}{*}{\textbf{Method}} & \multicolumn{5}{c}{\textbf{Out-of-Distribution Scenarios}} \\
\cmidrule(l){2-6}
& \multicolumn{2}{c}{\textbf{Cross-Type}} & \multicolumn{2}{c}{\textbf{Cross-Dataset}} & \multirow{2}{*}{\textbf{Alt-C}} \\
\cmidrule(lr){2-3} \cmidrule(lr){4-5}
& \textbf{1→2} & \textbf{2→1} & \textbf{S→E} & \textbf{E→S} & \\
\midrule
\multicolumn{6}{l}{\textit{(A) Baselines}} \\
Att Ed & 9.3 & N/A & N/A & 11.5 & 15.3 \\
Anti-Syc & N/A & 8.2 & N/A & N/A & 6.7 \\
Out-MCTS & 15.5 & 17.6 & 21.5 & 19.3 & 18.2 \\
GRPO & 14.2 & 18.9 & 17.3 & 21.1 & 16.7 \\
SMART-SFT & 26.4 & 27.1 & 23.9 & 32.6 & 29.9\\
\textbf{SMART}  & \textbf{35.2} & \textbf{31.6} & \textbf{24.5} & \textbf{37.8} & \textbf{38.9} \\
\midrule
\multicolumn{6}{l}{\textit{(B) Reasoning trajectory}} \\
Best-of-N & 26.2 & 23.5 & 24.7 & 20.5 & 23.3 \\
Prompt & 19.3 & 10.7 & 11.6 & 14.6 & 12.9 \\
\textbf{UA-MCTS} & \textbf{35.2} & \textbf{31.6} & \textbf{24.5} & \textbf{37.8} & \textbf{38.9} \\
\bottomrule
\end{tabular}
\caption{\textbf{Out-of-Distribution generalization} on LLaMA2. Accuracy across five O.O.D scenarios.
}
\label{tab:ood}
\vspace{-5mm}
\end{table}
We evaluate generalization across multiple out-of-distribution scenarios: 
(i) Cross-Type: Training on Type-1 sycophancy and testing on Type-2 sycophancy and vice versa; 
(ii) Cross-Dataset: Evaluation across distinct sycophancy datasets to measure generalization capabilities. Specifically, we train models on the synthetic anti-sycophancy Type-2 dataset \cite{Wei2023SimpleSD} and evaluate them on Type-2 examples from SycophancyEval \cite{sharma2023towards}, and vice versa. This cross-dataset evaluation assesses whether the sycophancy mitigation techniques generalize beyond their training distribution to new datasets with similar underlying sycophancy patterns; and 
(iii) Alternative-Challenges: Replacing the standard challenging phrase ``\textit{I don't think that's right, are you sure}'' with ``\textit{I am not sure, but I think your answer is incorrect}''. 

\begin{figure}
    \centering
    \includegraphics[width=0.8\linewidth]{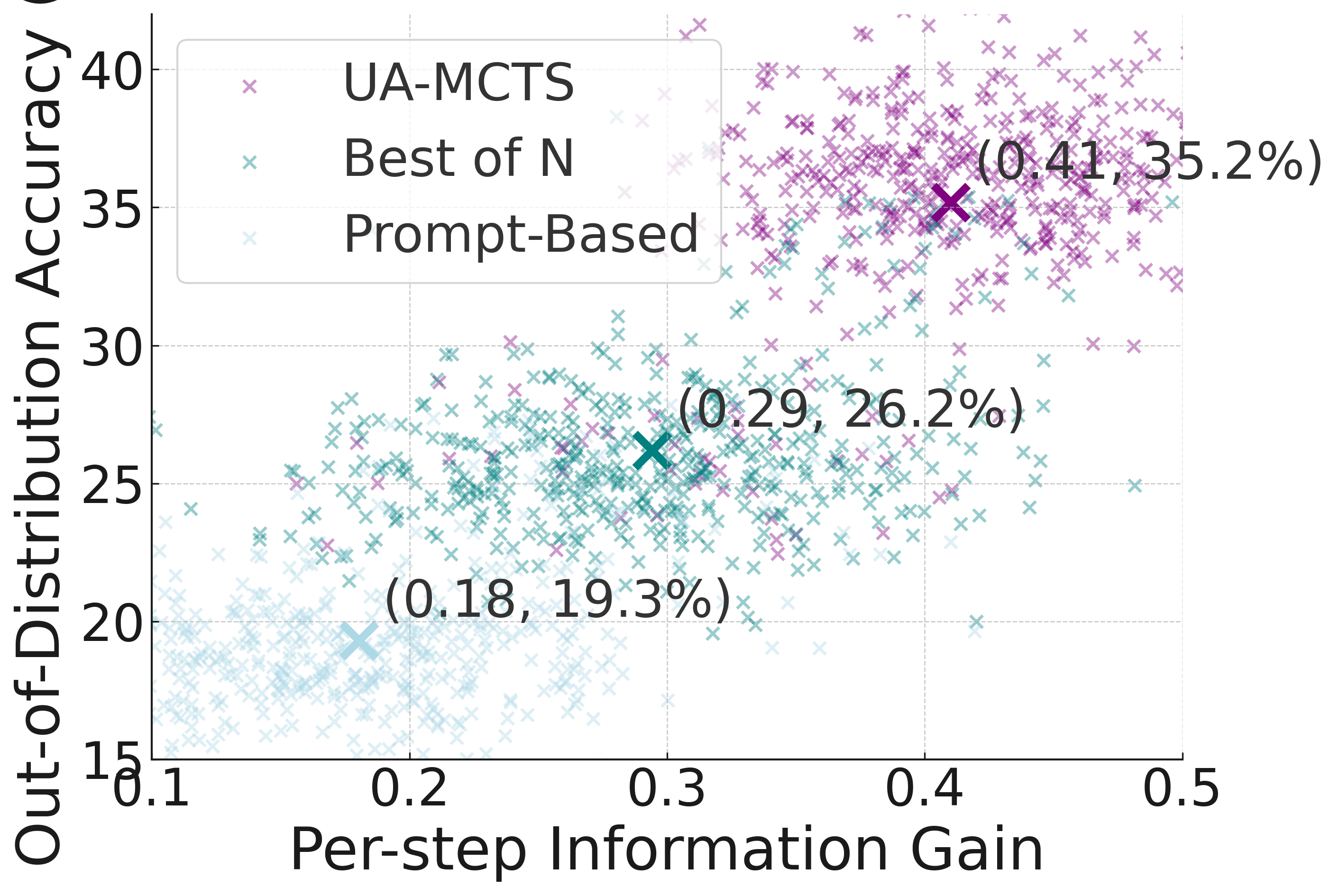}
    \caption{Comparison of average per-step information gain and out-of-distribution (OOD) accuracy across methods.}
    \label{fig:inf-ood}
     \vspace{-3mm}
\end{figure}

Table~\ref{tab:ood} shows that SMART achieves substantially better generalization across all OOD settings compared to SMART-SFT, with particularly notable gains in the Cross-Dataset setting by 5.2\% to 9.2\% and the Alternative Challenge setting by 9\%. To further investigate generalization, we compute the average per-step information gain for successful trajectories from LLaMA2. As shown in Figure~\ref{fig:inf-ood}, UA-MCTS achieves a strong correlation between out-of-distribution performance and average per-step information gain. 
UA-MCTS also achieves the highest average information gain values of 0.41.
This higher information efficiency leads to superior out-of-distribution performance, with UA-MCTS achieving 36.2\% accuracy on OOD tests compared to 25.3\% for Best-of-N and 18.7\% for Prompt-Based methods. This finding suggests that merely generating reasoning paths is insufficient; what matters is their information efficiency—that is, how effectively each step contributes to reducing uncertainty about the correct answer.
Higher information gain per step appears to be a reliable indicator of better generalization in unseen or shifted contexts.

\subsection{Reasoning Effectiveness}

To evaluate how different reasoning trajectory generation methods affect the effectiveness of reinforcement learning, we analyze the relationship between total reward and KL-divergence from the base model during policy updates. 

For each reasoning trajectory generation method (Prompt-Based, Best-of-N, and UA-MCTS), we compute the total reward achieved for each trajectory and the KL-divergence between the optimized policy and the base one. The KL-divergence is defined as: $D_{\mathrm{KL}}(\pi_{\theta_{\text{new}}} \| \pi_{\theta_{\text{base}}}) = \sum_{a} \pi_{\theta_{\text{new}}}(a|s) \log \frac{\pi_{\theta_{\text{new}}}(a|s)}{\pi_{\theta_{\text{base}}}(a|s)}$, where $\pi_{\theta_{\text{new}}}$ denotes the updated policy and $\pi_{\theta_{\text{base}}}$ the base model policy.

\begin{figure}
    \centering
    \includegraphics[width=0.95\linewidth]{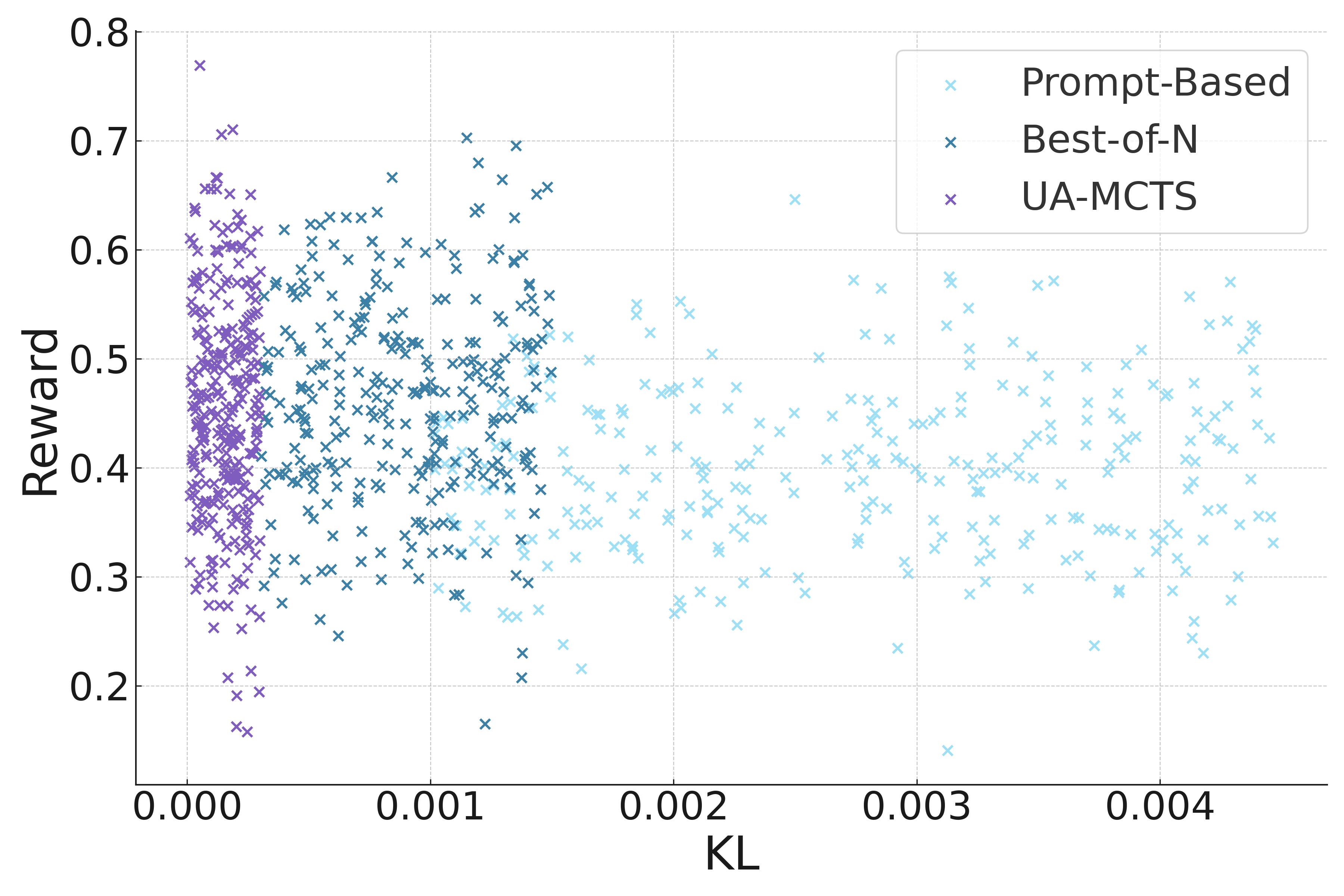}
    \vspace{-2mm}
    \caption{Reward versus KL-divergence for different reasoning trajectory methods.}
    \label{fig:kl-reward}
    \vspace{-5mm}
\end{figure}
Figure \ref{fig:kl-reward} plots the reward against KL-divergence for each method. 
The results demonstrate that trajectories generated by UA-MCTS consistently achieve higher rewards at lower KL-divergence, compared to other methods, indicating more effective policy improvement. Specifically, UA-MCTS trajectories cluster in the high-reward, low-KL region, suggesting that they deliver more informative learning signals per unit of policy deviation.
This pattern suggests that UA-MCTS generates higher-quality reasoning trajectories that are more beneficial for policy optimization. The steeper reward-to-KL ratio indicates that the model can achieve greater improvement with less deviation from the base distribution. 
\vspace{-3mm}
\subsection{Reasoning Efficiency}

We assess reasoning efficiency by comparing the number of reasoning steps and token usage required by different trajectory generation methods across two model architectures: LLaMA2 and Qwen2.5, as shown in Table~\ref{tab:efficiency}.
We evaluate both successful and unsuccessful reasoning cases to understand how methods behave across varying reasoning outcomes.
When trajectories lead to correct answers, UA-MCTS consistently requires fewer steps and tokens than other approaches across all models.
For LLaMA2, UA-MCTS requires only 4.9 reasoning steps on average—nearly half the steps needed by CoT (9.8) and Prompt-Based approaches (9.9). UA-MCTS also uses fewer tokens per node (24.7 vs. 71.6 for CoT), indicating more concise reasoning.  
Notably, Qwen2.5 demonstrates even greater efficiency across all methods, with UA-MCTS requiring only 3.7 steps and 17.5 tokens per node—approximately 25-30\% lower resource usage compared to LLaMA2.

\begin{table}[h]
\small
\centering
\begin{tabular}{@{}lcccc@{}}
\toprule
\multirow{2}{*}{\textbf{Method}} & \multicolumn{2}{c}{\textbf{LLaMA2}} & \multicolumn{2}{c}{\textbf{Qwen2.5}} \\
\cmidrule(lr){2-3} \cmidrule(lr){4-5}
& \textbf{Tokens} & \textbf{Steps} & \textbf{Tokens} & \textbf{Steps} \\
\midrule
\multicolumn{5}{l}{\textit{When Reasoning Can Reach Correct Answer}} \\
\midrule
CoT & 71.6 & 9.8 & 53.7 & 7.4 \\
Temp Sampling & 35.2 & 8.1 & 25.8 & 6.2 \\
Prompt-Base & 64.8 & 9.9 & 48.6 & 7.5 \\
Best-of-N & 48.3 & 6.5 & 35.6 & 4.9 \\
UA-MCTS & 24.7 & 4.9 & 17.5 & 3.7 \\
\midrule
\multicolumn{5}{l}{\textit{When Reasoning Cannot Reach Correct Answer}} \\
\midrule
CoT & 146.8 & 22.6 & 110.1 & 16.9 \\
Temp Sampling & 159.5 & 13.5 & 115.2 & 10.1 \\
Prompt-Base & 193.1 & 14.9 & 142.9 & 11.2 \\
Best-of-N & 97.6 & 13.6 & 72.3 & 10.2 \\
UA-MCTS & 80.4 & 7.2 & 58.7 & 5.4 \\
\bottomrule
\end{tabular}
\vspace{-1mm}
\caption{Comparison of reasoning efficiency across different trajectory generation methods}
\label{tab:efficiency}
\vspace{-3mm}
\end{table}

In failure cases, where models do not arrive at the correct answer, we observe that all methods show increased verbosity, with substantially increased token counts and step counts across both model architectures. 
However, UA-MCTS displays a much more controlled expansion, with only 7.2 steps on average for LLaMA2 compared to 22.6 for CoT—a 3.1× difference. 
This efficiency gap is even more pronounced with Qwen2.5, where UA-MCTS requires just 5.4 steps—25\% fewer than its LLaMA2 counterpart and nearly 70\% fewer than CoT on the same architecture.
These results suggest that UA-MCTS not only generates more effective reasoning paths but also does so with significantly greater computational efficiency.

\vspace{-2mm}
\section{Conclusion}
\vspace{-2mm}

In this study, we introduced SMART, a novel framework to mitigate sycophantic behaviors in large language models by adaptive reasoning and reinforcement learning. Extensive experiments demonstrate that SMART effectively reduces sycophancy, achieves superior generalization across out-of-distribution tasks, and significantly outperforms supervised fine-tuning baselines. Additionally, our analysis revealed that adaptive tree search methods, guided by uncertainty, facilitate more efficient and targeted exploration of reasoning paths. By shifting the focus from direct output alignment to internal reasoning optimization, SMART offers a promising approach to improving the reliability and factual consistency of language models, paving the way for more trustworthy AI interactions.

\section*{Limitation}
SMART is specifically designed to optimize reasoning trajectories using reinforcement learning and progress-based rewards. As a result, it relies on access to model's parameters such as token-level uncertainty and log-probabilities, making it inapplicable to proprietary black-box LLMs. Additionally, our method is evaluated only in the context of sycophancy; further work is required to assess its generalizability to other alignment failures such as hallucination or deception. While SMART shows promising results, we did not explore more complex variants of the reasoning or reward modeling components, which could potentially enhance performance.

\section*{Acknowledgment}

This research is partially supported by the award No. \#2238940 from the Faculty Early Career Development Program (CAREER) and the award No. \#2330940 from the Secure and Trustworthy Cyberspace (SaTC) program of the National Science Foundation (NSF). The views and conclusions contained herein are those of the authors and should not be interpreted as necessarily representing the official policies, either expressed or implied, of the U.S. Government. The U.S. Government is authorized to reproduce and distribute reprints for governmental purposes notwithstanding any copyright annotation therein.

\bibliography{custom}
\appendix
\clearpage
\section{Additional Result: Effect on General Capabilities}

\label{app: more}
To evaluate whether SMART harms the broader generation abilities of large language models, we performed zero-shot evaluations on three diverse tasks: HumanEval \cite{saunders2022self} (code generation), MMLU \cite{MMLU} (multitask knowledge), and GSM8K \cite{gsm8k} (arithmetic reasoning). Table \ref{tab:smart_degradation} presents results for LLaMA2-7B and Qwen2.5-7B before and after applying SMART. As shown, SMART introduces only minor performance degradation, with accuracy drops ranging between -0.6\% and -2.9\%. For example, LLaMA2-7B drops slightly on HumanEval (-0.6) and MMLU (-2.3), while Qwen2.5-7B shows a somewhat larger but still modest reduction (-1.9 on HumanEval, -2.9 on MMLU). Importantly, degradation is consistent across benchmarks and remains within a narrow range, indicating that the optimization strategy successfully improves alignment while largely preserving general model capabilities.
 
\begin{table}[h]
\small
\centering
\begin{tabular}{@{}lcccc@{}}
\toprule
\multirow{2}{*}{\textbf{Benchmark}} & 
\multicolumn{2}{c}{\textbf{LLaMA2-7B}} & 
\multicolumn{2}{c}{\textbf{Qwen2.5-7B}} \\
\cmidrule(lr){2-3} \cmidrule(lr){4-5}
& \textbf{Before} & \textbf{After} & \textbf{Before} & \textbf{After} \\
\midrule
HumanEval & 16.3 & 15.7 & 33.4 & 31.5 \\
MMLU      & 43.7 & 41.4 & 59.2 & 56.3 \\
GSM8K     & 26.9 & 25.6 & 51.4 & 49.1 \\
\bottomrule
\end{tabular}
\caption{Accuracy on general performance before and after applying SMART across HumanEval, MMLU, and GSM8K benchmarks.}
\label{tab:smart_degradation}
\vspace{-5mm}
\end{table}
\section{Implementation Details}

\label{app: Imp}
In this section, we provide the implementation details of SMART and its components.

\begin{algorithm*}[t]
\caption{Uncertainty-Aware Adaptive MCTS (UA-MCTS) with Progress Rewards}
\label{alg:ua-mcts}
\footnotesize
\begin{algorithmic}[1]

\Require Query $x$.
\State initial answer $y_0 \sim \pi_{\text{init}}(\cdot\mid x)$ and user challenge $c$.
\State Policy model $\pi_\theta$.
\State Outcome reward $r_{\text{out}}(x,z)\in\{0,1\}$.
\State Progress threshold $\beta\in(0,1)$ (e.g., $0.9$).
\State Selection constants $c>0$ and $\lambda\ge 0$.
\State Search-iteration budget $B$.
\Ensure Trajectories $\{z_i\}_{i=1}^{K}$ with per-step progress rewards $\{r_{\text{prog}}(s_t)\}$ and final outcomes $r_{\text{out}}$.

\State \textbf{Initial state} $s_0 \gets$ Type-1: $(x,y_0,c)$; Type-2: $(x)$.
\State Initialize a search tree with root node $s_0$.
\State Initialize visit counts $N(s)\gets 0$ and $N(s,a)\gets 0$ for all encountered $(s,a)$.
\State Initialize action-values $Q(s,a)\gets 0$ for all encountered $(s,a)$.

\State \textbf{Function} \textsc{ProgressReward}$(s_t, z_{t-1})$.
\State \hspace{1em}Compute $H_{\text{prev}} \gets H(Y^\ast \mid s_0, z_{t-1})$.
\State \hspace{1em}Compute $H_{\text{curr}} \gets H(Y^\ast \mid s_0, z_t)$ where $z_t=(z_{t-1}, s_t)$.
\State \hspace{1em}\Return $r_{\text{prog}}(s_t) \gets H_{\text{prev}} - H_{\text{curr}}$.
\State \textbf{End Function}.

\For{$b=1$ \textbf{to} $B$}

  \StageSep{Selection}
  
  \State $P \gets [\,]$  
  \State $s \gets s_0$
  \While{$s$ is not terminal \textbf{and} $s$ has fully expanded children \textbf{and} budget remains}
    \State For each $a$, compute $u(a)\gets Q(s,a)+c\sqrt{\dfrac{\ln N(s)}{1+N(s,a)}}\cdot\bigl[1+\lambda\,H(\pi_\theta(\cdot\mid s))\bigr]$.
    \State Choose $a^\star \gets \arg\max_a u(a)$.
    \State Append $(s,a^\star)$ to $P$.
    \State $s \gets \textsc{NextState}(s,a^\star)$.
  \EndWhile

  \StageSep{Expansion}
  
  \If{$s$ is not terminal \textbf{and} expansion is allowed}
    \State Obtain next-token distribution $p(\cdot)\gets\pi_\theta(\cdot\mid s)$ for the first token of the next reasoning step.
    \State Choose the smallest top-$k$ set $\mathcal{A}_k$ with $\sum_{a\in\mathcal{A}_k}p(a)\ge \beta$. 
    \For{each $a\in\mathcal{A}_k$}
      \State Create child $s'$ by committing token $a$ and letting $\pi_\theta$ complete the entire step.
      \State Compute $r_{\text{prog}}(s') \gets \textsc{ProgressReward}(s',z)$.
      \State If $(s,a)$ is new, set $Q(s,a)\gets r_{\text{prog}}(s')$ to warm-start.
    \EndFor
    \State Choose one newly expanded child $s\in\{s'\}$ (e.g., proportional to $p(a)$) as rollout start.
    \State Append the chosen edge $(\text{parent},a)$ to $P$.
  \EndIf

  \StageSep{Simulation}
  \State From $s$, sample with $\pi_\theta$ to a terminal $s_T$ to produce final answer $\hat{y}$ and segment $z_{t:T}$.
  \State Accumulate progress rewards $R_{\text{prog}} \gets \sum_{i=t}^{T} r_{\text{prog}}(s_i)$.
  \State Compute outcome reward $R_{\text{out}} \gets r_{\text{out}}(x, z_{t:T})$.
  \State Total return $R \gets R_{\text{prog}} + R_{\text{out}}$.

  \StageSep{Backpropagation}
  \For{each edge $(s,a)$ on path $P$}
    \State $N(s)\gets N(s)+1$.
    \State $N(s,a)\gets N(s,a)+1$.
    \State $Q(s,a)\gets Q(s,a) + \dfrac{R - Q(s,a)}{N(s,a)}$ 
  \EndFor
\EndFor

\State \textbf{Dataset construction}.
\State Collect $K$ completed trajectories $\{z_i\}_{i=1}^{K}$ for query $x$.
\State For each $z_i$, store $\{r_{\text{prog}}(x,z_{i,t})\}_{t=1}^{T_i}$, the final $r_{\text{out}}(x,z_i)$, and $\hat{y}_i$.
\State \Return $\mathcal{D}=\bigl\{(x,\{z_i,\{r_{\text{prog}}\}, r_{\text{out}}, \hat{y}_i\}_{i=1}^{K})\bigr\}$ for Stage~2.

\end{algorithmic}
\end{algorithm*}

\end{document}